\documentclass[letterpaper]{article} 
\usepackage{aaai25}  
\usepackage{times}  
\usepackage{helvet}  
\usepackage{courier}  
\usepackage[hyphens]{url}  
\usepackage{graphicx} 
\usepackage{natbib}  
\usepackage{caption} 
\usepackage{algorithm}
\usepackage{algorithmic}
\usepackage{amsmath}
\usepackage{amssymb}
\usepackage{booktabs}  
\usepackage[utf8]{inputenc}

\usepackage{newfloat}
\usepackage{listings}
\DeclareCaptionStyle{ruled}{labelfont=normalfont,labelsep=colon,strut=off} 
\floatstyle{ruled}
\newfloat{listing}{tb}{lst}{}
\floatname{listing}{Listing}
\title{Mice to Machines: Neural Representations from \\ Visual Cortex for Domain Generalization}
\author {
    Ahmed Qazi\thanks{Equal contribution.}\quad
    Hamd Jalil\footnotemark[1]\quad
    Asim Iqbal\thanks{Corresponding author.}
}
\affiliations {
    \normalfont
    \Large
    Tibbling Technologies\\
    \tt\small asim@tibbtech.com
}

\begin{document}

\maketitle

\begin{abstract}
The mouse is one of the most studied animal models in the field of systems neuroscience. Understanding the generalized patterns and decoding the neural representations that are evoked by the diverse range of natural scene stimuli in the mouse visual cortex is one of the key quests in computational vision. In recent years, significant parallels have been drawn between the primate visual cortex and hierarchical deep neural networks. However, their generalized efficacy in understanding mouse vision has been limited. In this study, we investigate the functional alignment between the mouse visual cortex and deep learning models for object classification tasks. We first introduce a generalized representational learning strategy that uncovers a striking resemblance between the functional mapping of the mouse visual cortex and high-performing deep learning models on both top-down (population-level) and bottom-up (single cell-level) scenarios. Next, this representational similarity across the two systems is further enhanced by the addition of \textbf{Neu}ral \textbf{R}esponse \textbf{N}ormalization (\textbf{NeuRN}) layer, inspired by the activation profile of excitatory and inhibitory neurons in the visual cortex. To test the performance effect of NeuRN on real-world tasks, we integrate it into deep learning models and observe significant improvements in their robustness against data shifts in domain generalization tasks. Our work proposes a novel framework for comparing the functional architecture of the mouse visual cortex with deep learning models. Our findings carry broad implications for the development of advanced AI models that draw inspiration from the mouse visual cortex, suggesting that these models serve as valuable tools for studying the neural representations of the mouse visual cortex and, as a result, enhancing their performance on real-world tasks. 
\end{abstract}

\section{Introduction}
\label{sec:intro}

Deep learning has revolutionized machine learning, enabling breakthroughs across a range of applications, particularly in computer vision \cite{11, 13, 14, 15, 16}. Deep neural networks (DNNs) have achieved remarkable performance in tasks such as image recognition, object detection, and semantic segmentation by learning complex, high-dimensional representations from large datasets.

Despite these advances, a critical challenge that persists is the problem of \textit{domain shift}, where models trained on a specific source domain perform poorly when applied to unseen target domains due to differences in data distributions \cite{blanchard2021domain, taori2020measuring, ben-david2010theory, recht2019imagenet, moreno-torres2012unifying}. This limitation hinders the deployment of DNNs in real-world scenarios, where data often vary widely and cannot be fully anticipated during training. Addressing domain shift requires models that can generalize across diverse environments by capturing domain-agnostic features robust to variations in input data. Biological visual systems, particularly the mouse visual cortex (MVC), excel at this kind of generalization, efficiently processing a wide range of visual stimuli with remarkable adaptability. The MVC's ability to extract invariant features from complex visual scenes offers valuable insights for improving domain generalization in artificial systems.

In this paper, we leverage principles from the MVC to tackle the domain shift problem in deep learning. Inspired from the Winner-Takes-All (WTA) mechanism observed in neuronal circuits \cite{Iqbal2024}, we applied this approach to enable \textbf{NeuRN}---\textbf{Neu}ral \textbf{R}esponse \textbf{N}ormalization---a biologically inspired technique designed to enhance both the domain generalization capabilities and biological alignment of DNNs. NeuRN normalizes pixel-level contrastive deviations to generate domain-agnostic feature representations, enabling generalization across varying domains. Furthermore, NeuRN increases the biological alignment between artificial neural networks and the MVC by incorporating properties of excitatory neurons into DNNs. Our contributions are:

\begin{itemize}
    \item \textbf{Framework for Biological and Artificial Representation Comparison:} We introduce a framework for comparing neural representations from the MVC and feature representations from DNNs, revealing a strong alignment, particularly with excitatory neurons. This comparison is conducted at both the population level and the single-neuron level, providing a comprehensive understanding of the similarities and differences between biological and artificial systems. {\textbf{Figure 1}} shows the block diagram of our comparison framework.

    \item \textbf{Development of NeuRN:} Taking inspiration from the computational principles of visual neurons \cite{Iqbal2024}, we applied it as a neuro-inspired normalization technique (NeuRN) that enhances domain generalization by capturing domain-agnostic features. By mimicking the response normalization property of excitatory and inhibitory neurons, NeuRN allows DNNs to focus on intrinsic structural features rather than superficial variations, improving their ability to generalize across different domains.

    \item \textbf{Evaluation of NeuRN's Impact:} We evaluate NeuRN's impact on both domain generalization tasks and biological alignment. Our experiments demonstrate that DNNs equipped with NeuRN not only perform better on challenging datasets such as MNIST, SVHN, USPS, and MNIST-M but also exhibit increased similarity to biological neural representations. We assess this alignment with and without NeuRN, showing that NeuRN enhances the biological plausibility of DNN representations.
\end{itemize}

By bridging the gap between biological and artificial neural systems, our work offers a novel approach to improving the robustness and adaptability of DNNs in the face of domain shift. Our findings have broad implications for the development of advanced AI models that draw inspiration from the visual processing capabilities of the brain, paving the way for more biologically plausible and generalizable artificial intelligence.

\begin{figure*}
    \centering
    \includegraphics[scale=0.7]{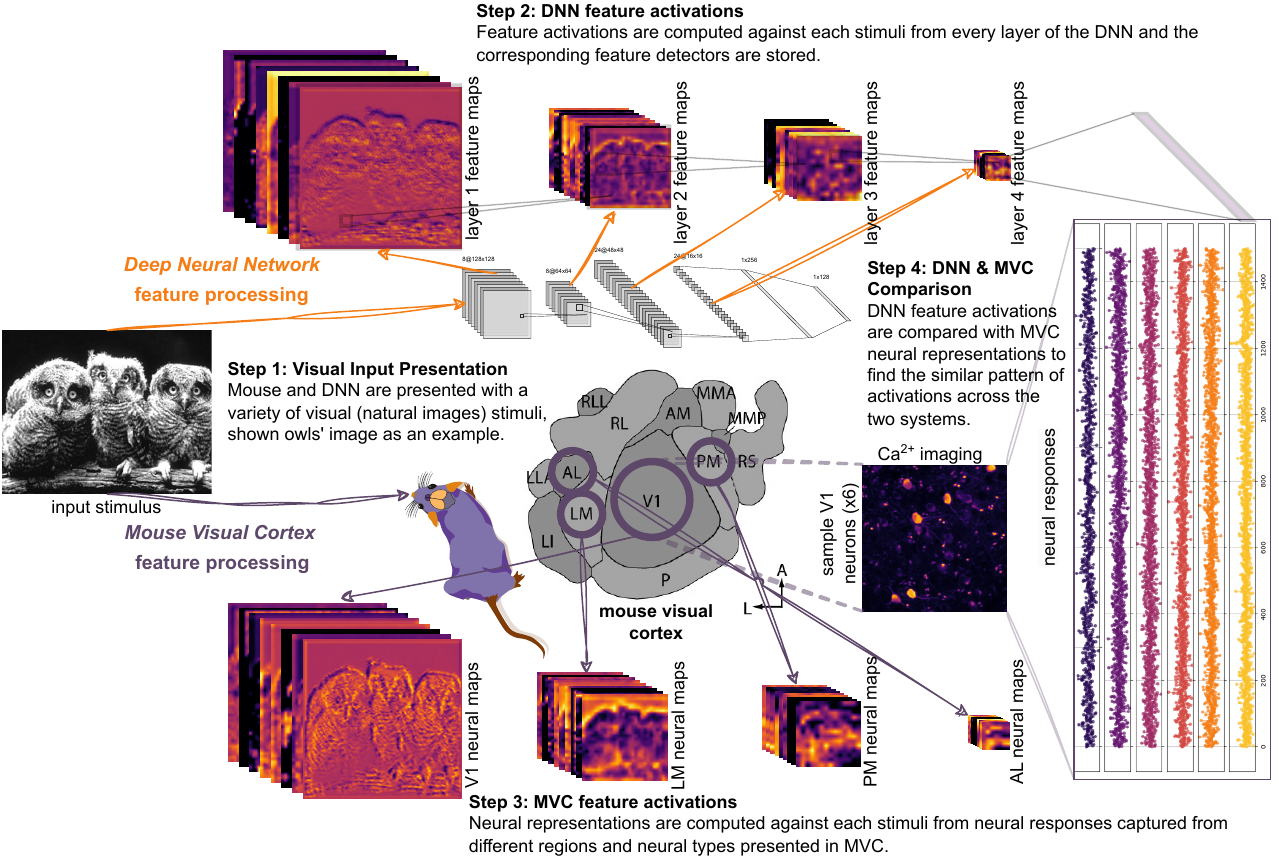}
    \caption{Block diagram architecture of comparison framework: image stimuli are presented to mice and DNN models (step 1) and their corresponding feature (step 2) and neural representations (step 3) are computed and systematically compared (step 4).} 
    \label{fig:1}
\end{figure*}

\section{Related Work}
\label{sec:related}
Neuro-inspired models have been widely studied to solve the problem of domain generalization. Convolutional Neural Networks (CNNs) have demonstrated the ability to predict neural responses in the MVC \cite{Cadena2019HowWD}. Comparing DNNs with biological neural responses has provided valuable insights into how artificial systems can emulate biological visual systems \cite{45, cvsnet2023, 49}. Furthermore, studies have shown that biasing CNN models toward a biological feature space results in more robust models \cite{50}. A significant body of work has focused on representational similarities between feature representations in DNNs and neural activity in response to various visual stimuli \cite{21, 22, 26, 27, 38, 49}. Models such as CORnet \cite{8,9} and frameworks like MouseNet \cite{45} and CVSNet \cite{cvsnet2023} have successfully incorporated biological principles, enabling DNNs to better mimic the efficiency and adaptability of the brain's visual processing \cite{veerabadranBioInspired}. For example, \cite{50} leveraged DNNs to probe the MVC, utilizing innovative regularization techniques to increase robustness against adversarial attacks.

While these studies have been pivotal in advancing our understanding of mammalian brain functionalities and inspiring DNNs, a common limitation is their scalability and applicability to diverse, real-world scenarios. Many proposed models, although theoretically profound, struggle to integrate effectively into practical applications across different domains. Building on this prior work, our framework compares MVC receptive fields (hereby referred to as neural representations) and DNN kernels (hereby referred to as feature representations), finding a strong similarity. This insight informs the design of NeuRN, which aligns DNN representations with biological plausibility, enabling robust generalization across varied visual tasks.

\section{Methods}

\subsection{Allen Brain Observatory dataset} 
To conduct the comparison between DNNs and MVC, we utilized the Allen Brain Observatory open-source dataset \cite{6}, which consists of in-vivo Calcium imaging data from GCaMP6-expressing neurons. These neurons were strategically selected from specific MVC areas, cortical layers, and Cre lines labeling a variety of neuron types. We specifically focused on the peak DF/F and time-to-peak characteristics. Furthermore, the mice in the experiment were presented with a variety of image stimuli, encompassing a wide spectrum of naturally occurring visual elements and environments ({{\textbf{Supplementary Figure S1}}), to closely mimic the diversity of visual inputs they might encounter in the wild. Neural activity during this image exposure was recorded and collated, providing a rich dataset on the response patterns of mouse visual neurons to different visual stimuli. Using these neural traces, we constructed the neural representations dataset to serve as the foundational basis for our comparative analyses.

\subsection{Neural Representations dataset}
\label{neural_reps}
Each stimulus from the Allen Brain Observatory natural scenes dataset \cite{6} was resized and flattened to create a uniform representation. This process resulted in a collection of flattened stimuli \(FS = \{fs_1, fs_2, \dots, fs_n\}\), where \(fs_i\) is the flattened representation of the \(i^{th}\) stimulus. For each neuron, we captured a series of trial traces in response to the stimulus presentations, forming \(T = \{t_1, t_2, \dots, t_n\}\), where each \(t_i\) is a concatenated vector of trial traces corresponding to the \(i^{th}\) stimulus presentation. The neural representation \(N\) for each neuron was computed as the product of the transpose of the trial trace matrix \(T^T\) and the flattened stimulus matrix \(FS\), resulting in:
\[
N = T^T \cdot FS
\]

Each element \( n_{ij} \) of the matrix \(N\) represents a 2D matrix that captures a detailed measure of the neuron's response, including both intensity and spatial distribution.

We analyzed neural responses from 1,000 neurons (500 excitatory and 500 inhibitory) selected using Uniform Manifold Approximation and Projection (UMAP) embeddings \cite{10} to capture the high-dimensional structure of the data. The UMAP parameters were set to \(n\_neighbors=15\), prioritizing local structure preservation, with a `euclidean' distance metric for similarity measurement, and \(min\_dist=0.1\) to prevent excessive clustering. Following UMAP dimensionality reduction, we applied K-means clustering with \(k=10\) to segment the neurons into clusters that balance granularity with over-segmentation. From each cluster, we extracted 50 central neural representations for both excitatory and inhibitory neurons, ensuring representative samples while avoiding outliers. This approach allowed us to obtain a diverse yet representative subset of neurons for our analysis.

\begin{figure*}
    \hspace{-5mm}
    \includegraphics[scale=0.8]{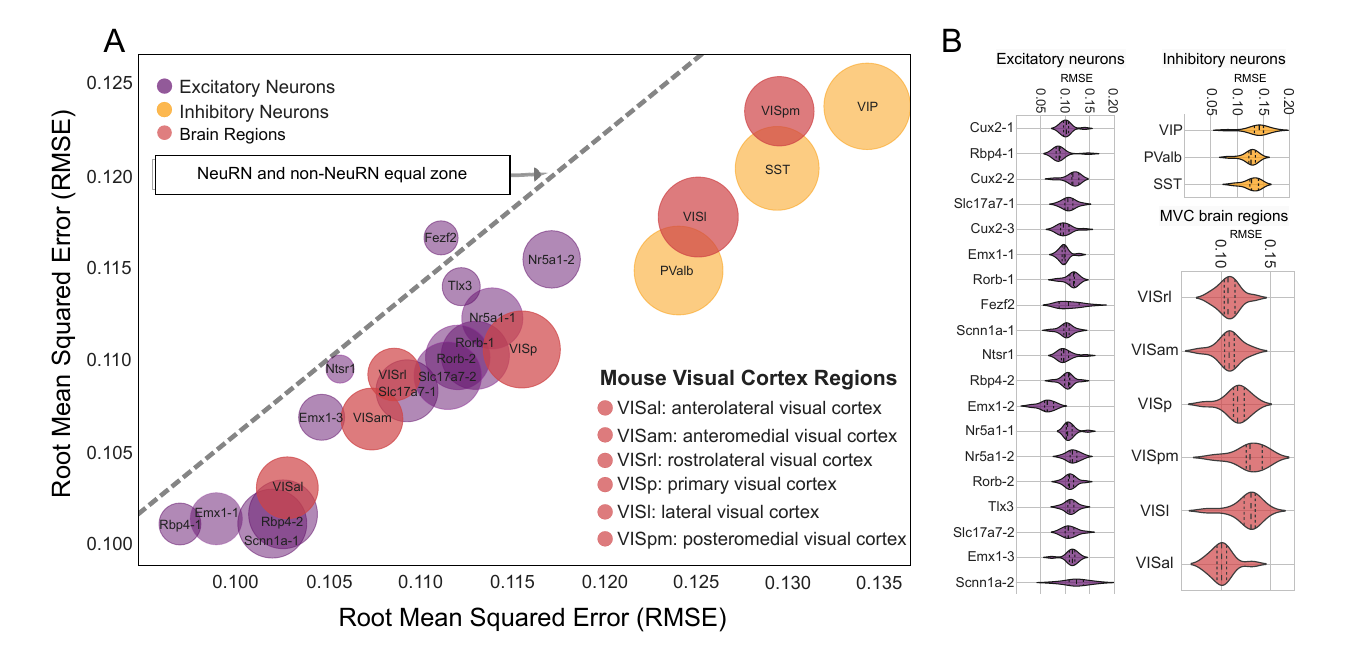}
    \caption{A) RMSE score comparison of feature representations from NeuRN and non-NeuRN DNN models with MVC neural representations. Data points below the diagonal line signify better NeuRN models' biological plausibility with MVC neural representations. B) RMSE scores of feature representations' comparisons with neural representations across neural types for 13 (non-NeuRN) DNN models.
    }
    \label{fig:5}
\end{figure*}

\subsection{Artificial feature representations}
To ensure that the DNNs have captured features from images similar to natural stimuli and are therefore comparable with neural representations from the MVC, we pretrained the DNNs on ImageNet and fine-tuned them on a specifically curated subset of CIFAR-100 (please refer to the supplementary materials for further details on DNN model training and the CIFAR-100 subset). To extract artificial feature representations, we utilized the convolutional layer \( L \) of the DNNs, defined by dimensions \([F,C,H,W]\), where \( F \) signifies the number of filters, \( C \) the number of channels, \( H \) the height, and \( W \) the width. We considered each channel of every filter individually and reshaped it to \([F \times C, H, W]\), interpreting each channel of a filter as a distinct 2D feature representation.

To select a representative sample of these feature representations for comparison with neural data, we applied UMAP embeddings to capture the underlying structure of the high-dimensional data. Subsequently, we performed K-means clustering (using the same parameters as for the biological neural representations) to group similar feature representations. This approach allowed us to generate a representative sample of 500 feature representations for each DNN, facilitating a meaningful comparison with the neural representations.

\subsection{High-level representational analysis}
To compare the neural and artificial feature representations, we considered an artificial feature representation \(K_1\) and a biological neural representation \(K_2\), both of size \(m \times n\). We reshaped \(K_1\) and \(K_2\) into 1D arrays:
\begin{align*}
    K1\_flat &= [k1_1, k1_2, \dots, k1_{mn}]  \\
    K2\_flat &= [k2_1, k2_2, \dots, k2_{mn}]
\end{align*}
With \(K1\_flat\) and \(K2\_flat\) mapped to an \(mn\)-dimensional space, we computed the Root Mean Squared Error (RMSE) between these vectors as a measure of dissimilarity:
\begin{equation}
    RMSE(K1\_flat, K2\_flat) = \sqrt{\frac{1}{mn} \sum_{i=1}^{mn} (k2_i - k1_i)^2}
\end{equation}

In our comparison methodology, which is essentially Representational Similarity Analysis (RSA), we use RMSE as the distance metric. This choice aligns with approaches in previous studies, such as those by \cite{khaligh2014deep}, where they discuss the use of Euclidean distance---closely related to RMSE---in RSA frameworks. We employ RMSE because it provides a direct and interpretable measure of the average magnitude of differences between corresponding elements in high-dimensional neural and model representations. RMSE captures both the direction and magnitude of discrepancies, making it sensitive to absolute differences in activation patterns. This sensitivity is crucial for our analysis, as we aim to assess not just the pattern similarity but also the fidelity of the model's predicted neural responses to the actual biological neural data. By using RMSE, we can quantify how closely our models replicate the neural representations, accounting for both pattern shape and activation magnitude, thereby providing a robust metric for evaluating representational alignment.

\begin{figure*}
    \hspace{-10mm}
    \includegraphics[scale=0.62]{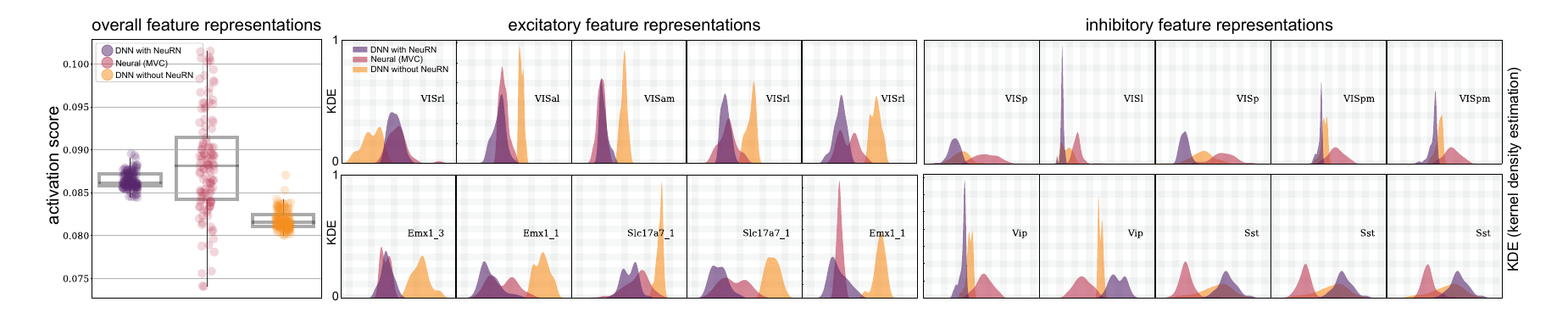}
    \caption{Distributional comparison of NeuRN and non-NeuRN feature representations with neural representations is shown. KDE curves of randomly sampled excitatory and inhibitory neurons show NeuRN augmented feature representations are more similar to biological representations.}
    \label{fig:6}
\end{figure*}

\subsection{Single neuron-level analysis}
Traditional efforts, such as those in \cite{23,49}, have aimed to understand the representational similarity of neural manifolds by mapping them to the architectural space of models. These approaches primarily focus on embedding high-level representations of the neural population, often overlooking the intricacies at the single neuron-level. Our work introduces a novel depth of exploration at this granular level, providing a richer understanding of the representational similarity between biological neural representations and artificial feature representations. In the MVC, the neural representations are used, captured by presenting specific stimuli. Analogously, in DNNs, feature representations act as the response to input stimuli. By comparing the feature representations from both NeuRN and non-NeuRN DNNs, we can contrast every neural representation in our dataset with the feature representations from each DNN layer, ensuring a comprehensive comparison with minimal data loss.

Our methodology offers a detailed examination of neural compatibility, going beyond the generalized shape metrics on neural representations explored in works like \cite{23}. While linear regression-based single neuron predictivity, as mentioned in \cite{49}, provides insights into neural representation correlations, our approach introduces a biophysical analysis that aligns feature representations at both the collective population and individual neuron levels. By emphasizing the distributional comparison of activations, we enrich traditional methodologies, extending beyond sole reliance on RSA. For each DNN layer, we retrieve its feature representations and evaluate their distributions using a Kernel Density Estimate (KDE). Similarly, we compute the KDE of the neural representations. The similarity between these KDEs is determined using the Intersection over Union (\(IoU\)) metric.

For each layer in a DNN, we extract the feature representations $x$ of $R$ dimensions and compute their distributions using a KDE:
\begin{equation}
K(u) = \frac{1}{\sqrt{2\pi} \sigma} \exp\left(-\frac{u^2}{2\sigma^2}\right)
\end{equation}
where $u$ is the distance from the center of the kernel, and the factor 
\begin{equation}
\frac{1}{\sqrt{2\pi} \sigma}
\end{equation}
ensures that the kernel integrates to 1.
The kernel density estimator for feature representations, ${f}(x)$, denoted as $M$:
\begin{equation}
\hat{f}(x) = \frac{1}{nh} \sum_{i=1}^{n} K\left(\frac{x-x_i}{h}\right)
\end{equation}
To get the KDE for 2D neural representations ($N$) we repeat the same process. We calculate the overlap between biological and artificial KDEs to quantify their similarity, where $IoU$ is derived as:
\begin{equation}
IoU = \frac{R \cap  N}{R \cup N}
\end{equation}

\subsection{CIFAR-100 subset: CIFAR-27}
Alongside our work with mice and natural scenes, we used models pre-trained on ImageNet, a large-scale and diverse image dataset widely used in deep learning. ImageNet pre-training enables the model to develop a broad set of feature detectors that can be applied across various visual tasks. We fine-tuned the models on a specifically curated subset of the CIFAR-100 dataset ({{\textbf{Supplementary Figure S2}}}). The CIFAR-100 dataset is a collection of 60,000 32x32 color images in 100 classes, with 600 images per class. Our subset comprised 27 classes, all pertaining to natural scenes or organisms. The classes included: butterfly, chimpanzee, bear, elephant, forest, fox, kangaroo, leopard, lion, lizard, maple\_tree, mountain, oak\_tree, orchid, otter, palm\_tree, pine\_tree, plain, poppy, rose, sea, sunflower, tiger, tulip, turtle, willow\_tree, and wolf. This data set was developed to ensure proximity to natural scenes stimuli. 


\subsection{Evaluation datasets}
Our domain generalization study utilizes diverse datasets including MNIST, SVHN, USPS, and MNIST-M, each presenting unique challenges \cite{1,2,3,4}. \textbf{MNIST} is a standard benchmark with 70,000 grayscale images of handwritten digits \cite{1}. \textbf{SVHN}, with over 600,000 images, features house numbers in real-world scenarios, adding complexity to digit recognition tasks \cite{2}. \textbf{USPS}, comprising 9,298 16x16 grayscale images from scanned mail \cite{3}. Finally, \textbf{MNIST-M} fuses MNIST digits with BSDS500's colored backgrounds, testing models against variations in color and texture \cite{4}. These datasets effectively test machine learning models' ability to generalize from clean, handwritten digits to recognizing numbers in varied, naturalistic settings. They serve as critical benchmarks to assess the generalization capabilities of vision models in diverse visual environments.

\section{Results \& Discussion}
In pursuit of structural clarity of this study, we outline our analysis as follows. We first extract neural representations from the MVC and artificial feature representations from DNNs to evaluate representational similarity. Our analysis reveal that excitatory neurons exhibit a significantly stronger correlation with artificial feature representations compared to inhibitory neurons. Based on these findings, we designed NeuRN by finetuning from \cite{Iqbal2024}, drawing inspiration from the response normalization property of neurons. Integrating NeuRN with DNNs led to a further increase in the similarity between artificial and biological neural representations. To assess NeuRN’s efficacy in improving generalization, we conduct domain generalization experiments using datasets such as MNIST, SVHN, USPS, and MNIST-M \cite{1,2,3,4}, which present varying challenges in digit recognition tasks. Our results demonstrate that NeuRN not only brings artificial feature representations closer to biological ones but also enhances DNNs' ability to capture domain-agnostic features, leading to improved generalization across different domains. The following sections detail each step of our analysis.

\subsection{Strong alignment between feature and neural representations}
Representational analysis was performed on neural representations and feature representations from the selected DNN models to quantify the similarity between biological and artificial responses to stimuli. This comparison established a benchmark for how closely the artificial models could mimic the biological neural representations. The analysis, shown in \textbf{Figure~\ref{fig:5}(B)}, indicates that the biological neural representations and artificial feature representations have a significant degree of similarity, as evidenced by their RMSE scores. This result reaffirms the connection between biological and artificial neural networks, which has been previously suggested in studies such as \cite{8,45,46,47,48,49}. We measured the similarity across all brain regions in the MVC, as well as across different neural genotypes. Among all the brain regions studied, the anterolateral visual area (VISal) prominently stood out for its close resemblance to DNN feature representations. This result aligns with the findings of Conwell et al. \cite{49}, which highlighted the distinct patterns in VISal that mirrored those captured by DNNs. Moreover, excitatory neurons consistently exhibited lower RMSE compared to inhibitory neurons (\textbf{Figure~\ref{fig:5}(B)}), emphasizing a stronger correlation between the activity of excitatory neurons and DNN feature representations.

These findings suggest that DNNs are more effective at capturing the representational characteristics of excitatory neurons in the MVC. This strong alignment supports the potential of leveraging biological insights to inform and improve artificial neural network designs.

\subsection{Neural Response Normalization (NeuRN)}
\label{sec:neurn}

\begin{table*}[htbp]
\centering

\scalebox{0.9}{
\label{tab:table}

\begin{tabular}{l|*{12}{c}}
\toprule
\textbf{Models} & \rotatebox{45}{\textbf{M$\rightarrow$U}} & \rotatebox{45}{\textbf{M$\rightarrow$S}} & \rotatebox{45}{\textbf{M$\rightarrow$MM}} & \rotatebox{45}{\textbf{U$\rightarrow$M}} & \rotatebox{45}{\textbf{U$\rightarrow$S}} & \rotatebox{45}{\textbf{U$\rightarrow$MM}} & \rotatebox{45}{\textbf{S$\rightarrow$M}} & \rotatebox{45}{\textbf{S$\rightarrow$U}} & \rotatebox{45}{\textbf{S$\rightarrow$MM}} & \rotatebox{45}{\textbf{MM$\rightarrow$M}} & \rotatebox{45}{\textbf{MM$\rightarrow$U}} & \rotatebox{45}{\textbf{MM$\rightarrow$S}} \\
\midrule

VGG19+NeuRN & 
\textbf{74.2} & 
\textbf{24.5} & 
\textbf{62.6} & 
48.4 & 
\textbf{20.4} & 
\textbf{30.5} & 
\textbf{51.3} & 
28.4 & 
\textbf{38.4} & 
\underline{\textbf{96.2}} & 
74.3 & 
\textbf{33.6} \\

VGG19 & 
70.9 & 
13.7 & 
39.7 & 
66.2 & 
13.9 & 
19.9 & 
47.2 & 
32.5 & 
33.9 & 
94.3 & 
76.7 & 
22.5 \\
\hline

EfficientNetB0+NeuRN & 
7.6 & 
6.8 & 
18.2 & 
\textbf{11.6} & 
\textbf{10.8} & 
\textbf{10.8} & 
24.2 & 
\textbf{14.9} & 
\textbf{21.1} & 
93.1 & 
\textbf{51.3} & 
\textbf{21.8} \\

EfficientNetB0 & 
8.6 & 
7.4 & 
18.4 & 
9.4 & 
8.0 & 
10.6 & 
34.6 & 
12.0 & 
18.6 & 
96.8 & 
17.8 & 
12.5 \\
\hline
DenseNet121+NeuRN & 
26.4 & 
\textbf{15.9} & 
\textbf{50.8} & 
\textbf{43.1} & 
\textbf{14.8} & 
\textbf{21.0} & 
\textbf{49.1} & 
20.0 & 
\textbf{32.6} & 
85.2 & 
54.0 & 
\textbf{26.3} \\

DenseNet121 & 
74.3 & 
11.7 & 
38.4 & 
39.5 & 
11.4 & 
20.9 & 
34.1 & 
22.1 & 
31.5 & 
96.6 & 
73.3 & 
21.3 \\
\hline

ShuffleNet+NeuRN & 
7.0 & 
\textbf{56.1} & 
\textbf{71.1} & 
22.6 & 
\textbf{36.2} & 
\textbf{27.2} & 
\textbf{32.1} & 
27.5 &
\textbf{28.9} & 
\underline{\textbf{84.8}} & 
\textbf{7.6} & 
\textbf{50.4} \\

ShuffleNet & 
9.3 & 
19.0 & 
14.1 & 
44.9 & 
12.2 & 
13.0 & 
26.1 & 
31.1 & 
19.7 & 
38.7 & 
3.2 & 
18.3 \\
\hline

Xception+NeuRN & 
61.9 & 
\textbf{20.9} & 
\textbf{59.7} & 
\textbf{41.3} & 
\textbf{15.9} & 
\textbf{25.8} & 
\textbf{48.8} & 
\textbf{16.4} & 
\textbf{37.7} & 
95.7 & 
31.8 & 
\textbf{33.1} \\

Xception & 
62.4 & 
19.4 & 
56.0 & 
20.9 & 
11.9 & 
21.1 & 
42.7 & 
14.2 & 
34.2 & 
97.9 & 
77.8 & 
25.6 \\
\hline

NASNetMobile+NeuRN & 
\textbf{45.4} & 
\textbf{21.1} & 
\textbf{47.5} & 
\textbf{25.5} & 
\textbf{10.4} & 
16.9 & 
\textbf{47.6} & 
16.5 & 
\textbf{35.1} & 
\underline{\textbf{95.6}} & 
58.3 & 
\textbf{31.1} \\

NASNetMobile & 
33.0 & 
12.7 & 
36.0 & 
24.8 & 
10.1 & 
21.4 & 
46.0 & 
23.8 & 
32.6 & 
90.0 & 
63.7 & 
20.3 \\
\hline



ResNet50v2+NeuRN & 
31.4 & 
\textbf{23.1} & 
\textbf{59.6} & 
\textbf{28.4} & 
\textbf{14.1} & 
\textbf{20.7} & 
\textbf{50.5} & 
\textbf{18.8} & 
\textbf{35.7} & 
96.4 & 
55.7 & 
\textbf{31.3} \\

ResNet50v2 & 
78.7 & 
21.6 & 
57.8 & 
11.1 & 
8.6 & 
10.5 & 
48.0 & 
15.6 & 
33.7 & 
97.2 & 
78.1 & 
17.2 \\
\hline

Inceptionv3+NeuRN & 
52.6 & 
\textbf{22.6} & 
\textbf{62.2} & 
\textbf{50.8} & 
\textbf{14.4} & 
\textbf{31.3} & 
47.9 & 
19.7 & 
37.6 & 
96.1 & 
75.5 & 
\textbf{34.0} \\

Inceptionv3 & 
69.8 & 
16.4 & 
57.8 & 
34.2 & 
8.9 & 
24.9 & 
53.5 & 
28.0 & 
39.0 & 
97.2 & 
82.0 & 
15.0 \\
\hline



MobileNetv2+NeuRN & 
\textbf{20.9} & 
\textbf{16.3} & 
45.0 & 
\textbf{34.2} & 
\textbf{10.6} & 
17.5 & 
\textbf{48.9} & 
15.0 & 
\textbf{36.2} & 
\underline{\textbf{96.2}} & 
\underline{\textbf{88.5}} & 
\textbf{33.9} \\

MobileNetv2 & 
8.8 & 
16.1 & 
45.9 & 
32.6 & 
9.3 & 
17.8 & 
45.9 & 
17.7 & 
30.8 & 
94.2 & 
44.9 & 
20.8 \\
\hline
ViT+NeuRN & 
61.9 & 
\textbf{16.2} & 
\textbf{44.7} & 
\textbf{36.6} & 
\textbf{14.1} & 
\textbf{25.3} &
33.2 & 
25.5 & 
26.5 & 
87.8 & 
58.9 & 
\textbf{23.5} \\

ViT & 
69.4 & 
13.1 & 
32.8 & 
26.2 & 
9.4 & 
15.6 & 
45.2 & 
36.5 & 
30.6 & 
95.5 & 
64.4 & 
21.8 \\
%





\bottomrule
\end{tabular}
}
\caption{The table compares the performance of image-classification DNNs across domain transfer tasks from the source to the target (source $\rightarrow$ target) domains. The domains involved are MNIST (M), SVHN (S), USPS (U), and MNIST-M (MM). Entries in bold indicate an enhancement in performance due to the NeuRN adaptation, and underlined values are close-to-benchmark scores.}

\end{table*}

Building on our observation of a strong similarity between excitatory neurons and DNN feature representations, and after incorporating insights from Winner-Takes-All (WTA) mechanism observed in neuronal circuits \cite{Iqbal2024}, we adapted Neural Response Normalization (NeuRN), a neuro-inspired layer that embeds the response normalization property of excitatory and inhibitory neurons into DNNs \cite{zayyad2023normalization, burg2021learning, sawada2016emulating, das2021effect}. These neurons are distinguished by their ability to encode the structure and contrast of input images. By incorporating this biological functionality into DNNs through NeuRN, we aim to enhance the models' ability to replicate the natural processing capabilities of the brain.

Given an input image denoted by \(I \in \mathbb{R}^{W \times H \times C}\), where \(W\), \(H\), and \(C\) represent the width, height, and channels of the image, respectively, we aim to obtain a domain-agnostic representation \(I_{a}\) of this image. The procedure involves extracting patches of size \(k\) from each channel in \(I\). Assuming a stride of \(1\), the total number of these patches equals the pixel count of the input. We represent the collection of these patches as \(P = \{p_{k_1}, p_{k_2}, \dots, p_{k_n}\}\), where \(p_k\) is defined as the patch of size \(k\) surrounding the pixel \(x\) located at coordinates \((i, j)\) within the image. We then compute the mean \(\mu_{p_k}\) and standard deviation \(\sigma_{p_k}\) of each patch:
\[
\mu_{p_k} = \frac{1}{k^2} \sum_{i,j \in p_k} x_{ij}, \quad \sigma_{p_k} = \sqrt{\frac{1}{k^2} \sum_{i,j \in p_k} (x_{ij} - \mu_{p_k})^2}
\]
Finally, we normalize the standard deviation to obtain the domain-agnostic representation:
\[
I_a = \frac{1}{c} \cdot \sigma_{p_k}, \quad \text{where } c = \max(\sigma)
\]

By using the standard deviation, \( \sigma_{p_k} \), NeuRN captures the contrast of local features, leading to representations that can generalize better across domains. Unlike Local Response Normalization (LRN) \cite{11}, which focuses on channel-wise activations using ratios of activations to neighbouring sums, and Local Contrast Normalization (LCN) \cite{Jarrett2009WhatIT}, which enhances individual pixel contrast but may miss the larger structural context, NeuRN considers spatial information of patches and maintains structural details throughout the image. This holistic, context-aware approach accommodates both local and global patterns, making NeuRN robust across varied visual settings. By aligning closely with neurobiological insights—specifically how neurons in the visual cortex encode contrasting features—NeuRN provides a more biologically plausible model for response normalization.

\subsection{NeuRN improves feature and neural representational alignment in DNNs}
Our results show that after incorporating NeuRN into the models and subjecting them to the same training regimen (pretrained on ImageNet and fine-tuned on a subset of CIFAR-100), models with NeuRN achieved lower RMSE scores compared to those without NeuRN. This demonstrates NeuRN’s effectiveness in enhancing the models’ ability to mimic biological neural representations. As shown in \textbf{Figure~\ref{fig:5}(A)}, all genotypes, except for Fezf2, exhibited lower RMSE scores in NeuRN models compared to their non-NeuRN counterparts. This suggests that NeuRN’s normalization effect improves representational similarity and, consequently, biological plausibility among DNN architectures. Furthermore, we observe that feature representations of models correspond more closely with excitatory neural representations, characterized by genotypes such as Rbp4, Emx1, and Scnn1a, than with inhibitory neural representations, associated with genotypes like Sst and Vip. However, the differences are insignificant. Such findings align with existing biological literature, which emphasizes the pivotal role of these biological neurons in information processing and signal transmission within neural circuits \cite{28}. The adaptation of NeuRN further accentuates this correlation.
\\
\indent
These results underscore the potential of NeuRN in bridging the gap between artificial and biological neural representations, enhancing the biological plausibility of DNNs, and potentially leading to models that more accurately reflect neural processing in the brain.

\subsection{NeuRN improves feature and neural representational alignment at Single neuron-level}
Our experiments demonstrate that NeuRN-activated feature representations exhibit a higher average activation mean compared to those without NeuRN. This increase in activation levels makes the model's neurons more comparable to excitatory neurons in the biological brain, known for their significant role in signal transmission and higher firing rates. As shown in \textbf{Figure~\ref{fig:6}}, there is substantial overlap between the activation profiles (measured using KDE) of excitatory genotypes and brain regions and those of NeuRN feature representations, indicating a close alignment with the distribution of biological neural responses. This similarity reflects not just activation patterns but also key statistical properties like mean activation levels and variability, which are crucial for accurate information processing in neural networks.

These findings align with our high-level representational analysis, reinforcing NeuRN's consistent impact across different analysis levels. By enhancing activation levels in DNNs to match those of excitatory neurons, NeuRN improves the models' ability to emulate the functional characteristics of biological neural networks. Thus, incorporating NeuRN increases the biological plausibility of DNN feature representations, particularly among excitatory neurons. This enhanced alignment suggests that NeuRN enables DNNs to better capture the intrinsic properties of biological neurons, leading to models that are more robust in domain generalization tasks and more reflective of actual neural processing in the brain.

\subsection{NeuRN-derived performance boost for domain generalization in DNNs}
Using the digit datasets MNIST, SVHN, USPS, and MNIST-M, we test NeuRN’s domain-agnostic feature encoding capabilities and evaluat its impact on domain generalization. These datasets present unique challenges due to variations in image quality, style, and background noise, making them suitable for testing domain generalization capabilities. Our results, summarized in \textbf{Table 1}, show that DNNs fine-tuned on a single source domain and tested on various target domains exhibited significant improvement in cross-domain evaluation when equipped with NeuRN. This improvement is particularly evident in transitions from MNIST-M to SVHN, MNIST to SVHN, MNIST to MNIST-M, and USPS to SVHN, highlighting NeuRN’s strong domain bridging capability and validates our previous findings \cite{Iqbal2024}. The significance of these results lies in NeuRN's ability to enhance the robustness of DNNs against domain shifts, a critical challenge in deploying machine learning models in real-world scenarios where data distributions often differ from training conditions. By normalizing input data in a biologically inspired manner, NeuRN enables models to focus on intrinsic structural features rather than superficial variations, leading to improved generalization across diverse datasets.

\section{Conclusion}
Our work presents a framework to compare biological neural and artificial feature representations, allowing for both population-level and single-neuron analysis across brain regions and genotypes from the MVC. Aligned with previous findings, this framework can be applied to any neuron subgroup for future research. We also adapted NeuRN from Winner-Takes-All (WTA) mechanism observed in neuronal circuits, a brain-inspired technique that enhances domain generalization, improving DNN performance on image classification tasks. NeuRN captures domain-agnostic representations in models like CNNs and Vision Transformers (ViTs), demonstrating its potential for broader applications. Ongoing experiments are exploring NeuRN’s effectiveness in downstream tasks, including segmentation, registration, and novel view synthesis. Our findings carry broad implications for the development of advanced AI models that draw inspiration from the mouse visual cortex. By integrating biological principles into DNNs, we can develop models that not only perform better on real-world tasks but also provide insights into the neural representations of the brain. This cross-disciplinary approach holds promise for both neuroscience and artificial intelligence, paving the way for more robust and biologically plausible AI systems.

\renewcommand{\figurename}{Supplementary Figure}
\renewcommand{\tablename}{Supplementary Table}

\setcounter{figure}{0}
\setcounter{table}{0}

\renewcommand{\thefigure}{S\arabic{figure}}
\renewcommand{\thetable}{S\arabic{table}}

\section{Appendix}
\vspace{5mm}
\begin{figure*}
    \centering
    \begin{minipage}{0.45\textwidth}
        \centering
        \includegraphics[width=1\textwidth]{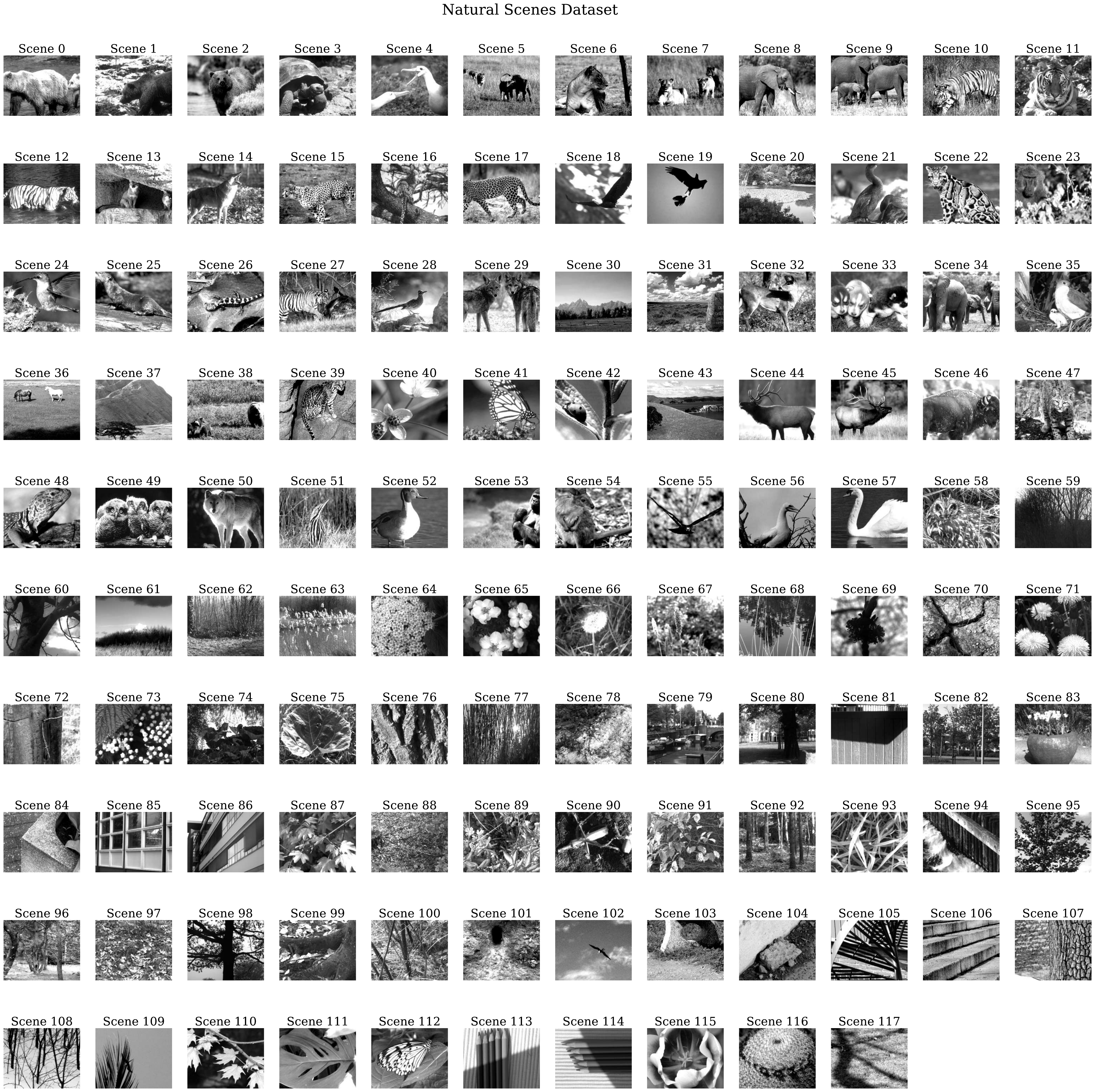}
        \caption{A comprehensive overview of the Natural Scenes Dataset, which comprises a diverse array of real-world images. Each image presents a unique scene, contributing to the dataset's broad scope that spans across various landscapes, urban areas, and natural phenomena.}
    \end{minipage}\hfill
    \begin{minipage}{0.45\textwidth}
        \centering
        \includegraphics[width=1\textwidth]{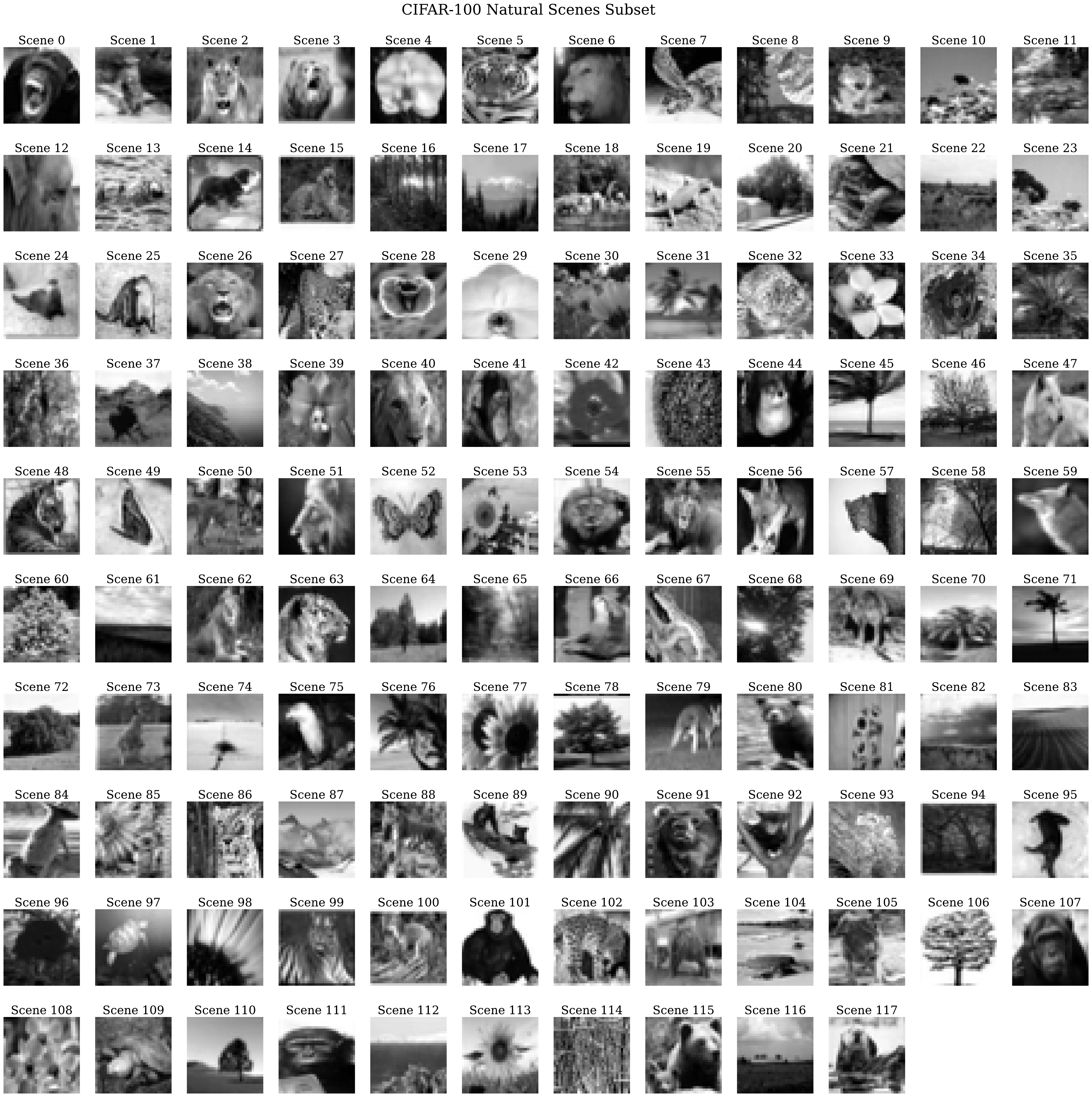} 
        \caption{A curated subset of the CIFAR-100 dataset, specifically focusing on natural scenes. These images cover a wide spectrum of natural environments, providing a compact yet varied dataset for studying natural image perception and related neural dynamics.}
    \end{minipage}
    \label{fig:ns}
\end{figure*}

\begin{figure*}
    \centering
    \includegraphics[scale=0.18]{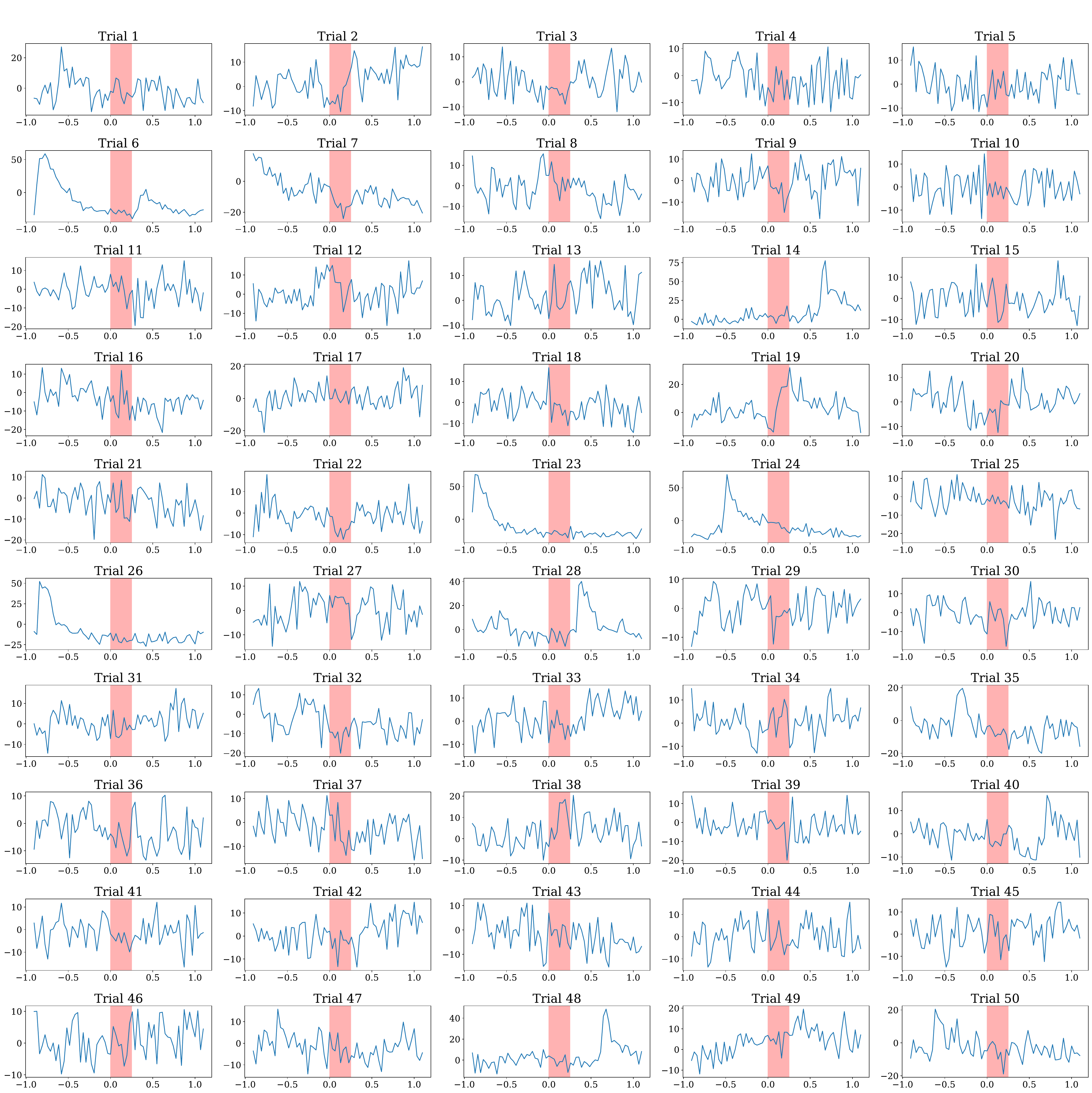}
    \caption{A variety of DF/F traces representing activation of different neurons for various trials, providing insights into their unique signature of activity profile. Here x-axis is time in seconds and y-axis is neural amplitude.}
    \label{fig:dff}
\end{figure*}

\begin{figure*}
    \centering
    \begin{minipage}{0.40\textwidth}
        \centering
        \includegraphics[width=1\textwidth]{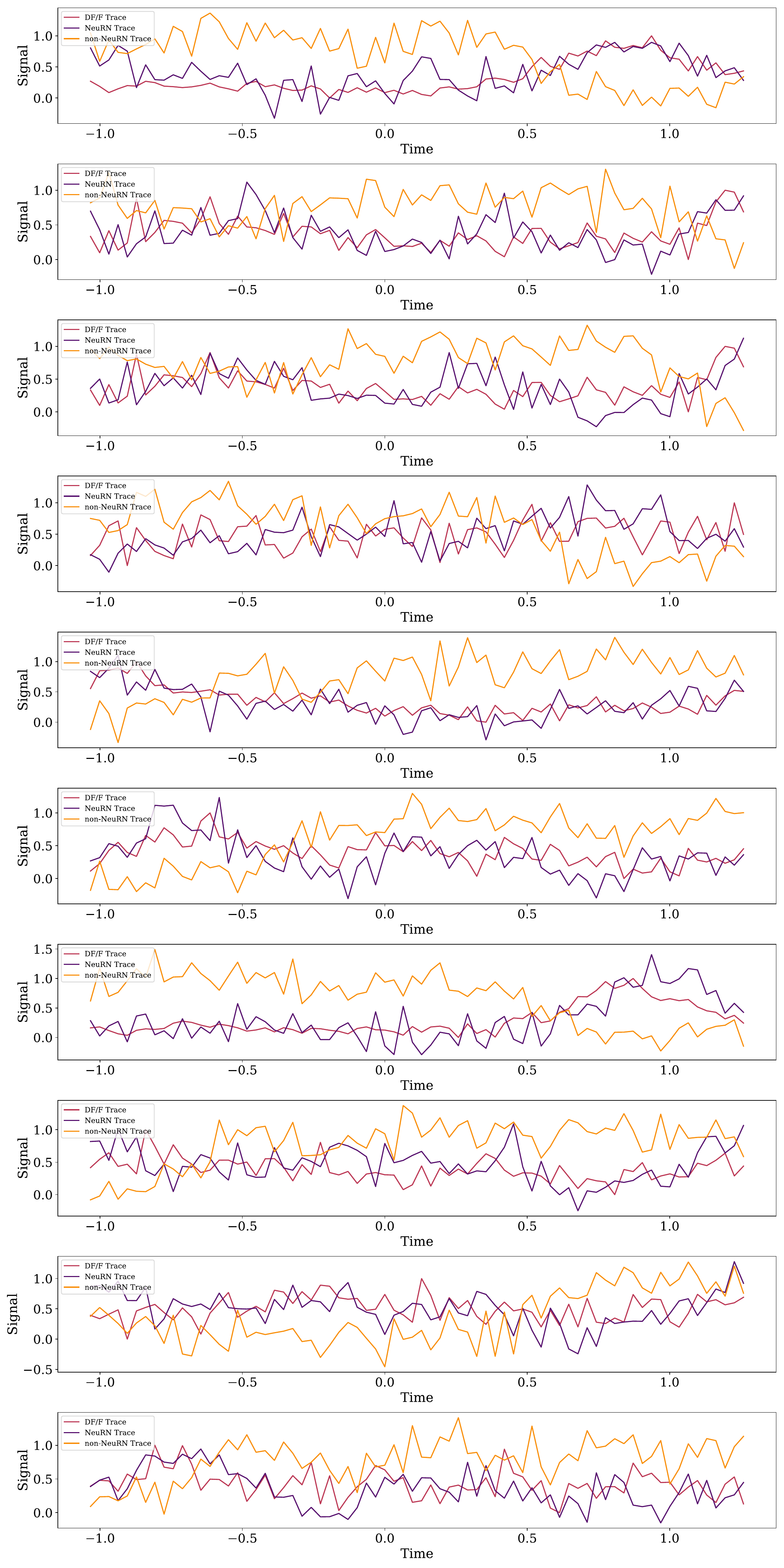} 
        \caption{Overlay of DF/F traces from brain region VISal and 1D feature representation signals of NeuRN and non-NeuRN models, highlighting the close alignment of NeuRN traces with DF/F traces in certain brain regions. Here x-axis is time in seconds and y-axis is neural amplitude.}
    \end{minipage}\hfill
    \begin{minipage}{0.40\textwidth}
        \centering
        \includegraphics[width=1\textwidth]{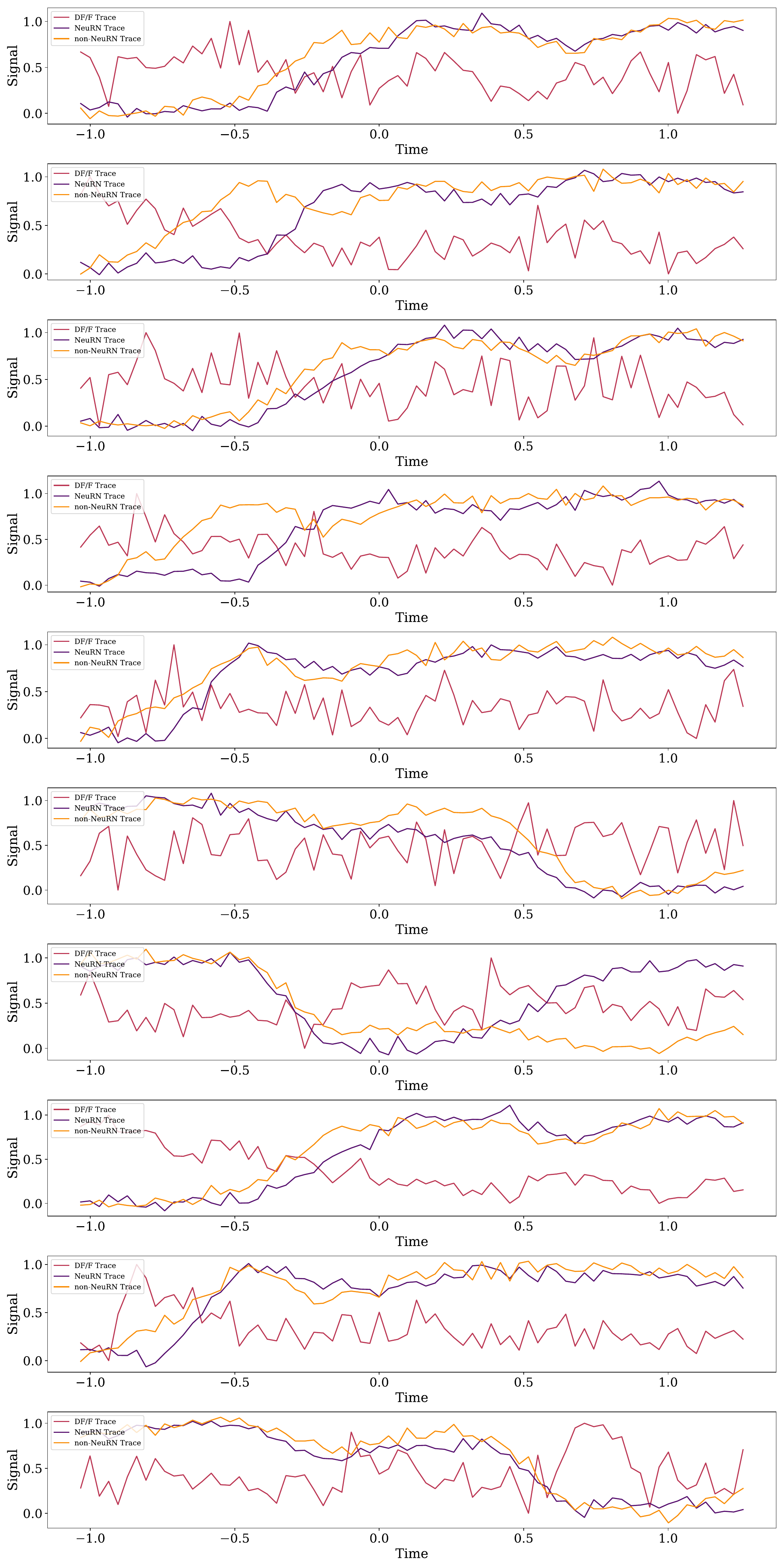} 
        \caption{Overlay of DF/F traces from brain region VISpm and 1D feature representation signals of NeuRN and non-NeuRN models, illustrating a case where NeuRN models struggle to accurately capture the neural dynamics. Here x-axis is time in seconds and y-axis is neural amplitude.}
    \end{minipage}
    \label{fig:traces}
\end{figure*}

\subsection{Neural responses}
{{\textbf{Supplementary Figure S3}}} presents a unique visualization of DF/F traces corresponding to different neurons. The plot provides a glimpse into the temporal dynamics of neuronal responses offering a detailed perspective of how these signals evolve over time, pre- and post- stimulus presentation, compared to the peak DF/F values. The variance observed in the profiles and magnitude of these DF/F traces indicates the diversity of neuronal response patterns based on factors such as brain region, genotype, and functional characteristics.

Furthermore, \textbf{Supplementary Figure S4} shows an overlay of DF/F traces from the VISal brain region alongside 1D feature representation signals from NeuRN and non-NeuRN models. It highlights the close alignment of NeuRN-based traces with the biological DF/F traces, demonstrating NeuRN’s capability to effectively capture neural patterns in this brain region. The x-axis represents time in seconds, while the y-axis corresponds to neural amplitude. In contrast, the \textbf{Supplementary Figure S5} presents an overlay of DF/F traces from the VISpm brain region with the same feature representation signals. This figure illustrates a case where NeuRN models struggle to fully replicate the neural dynamics of the region, as reflected in the less precise alignment compared to VISal. These observations underscore NeuRN’s strengths in certain brain regions while pointing to potential challenges in regions with complex neural dynamics.

\subsection{DNN training details}
We used DNN models pre-trained on ImageNet, a large-scale and diverse image dataset widely used in deep learning. ImageNet pre-training enables the model to develop a broad set of feature detectors that can be applied across various visual tasks. We fine-tuned the models on a specifically curated subset of the CIFAR-100 dataset as described. We applied these models to the MNIST datasets for domain generalization tasks. Throughout the training phase, early stopping was consistently applied to prevent over-fitting and to optimize performance. In order to train the DNN models, a learning rate of 0.001, batch size of 256 and Adam optimizer was used. Early stopping patience of 5 was used with validation accuracy as a performance metric. All experiments were performed using NVIDIA TITAN RTX GPU.

\end{document}